# Improved Multi-Class Cost-Sensitive Boosting via Estimation of the Minimum-Risk Class

**Ron Appel**
Caltech
appel@caltech.edu

**Xavier Burgos-Artizzu**
Transmural Biotech
Caltech
xpburgos@transmuralbiotech.com

**Pietro Perona**
Caltech
perona@caltech.edu

## Abstract

We present a simple unified framework for multi-class cost-sensitive boosting. The minimum-risk class is estimated directly, rather than via an approximation of the posterior distribution. Our method jointly optimizes binary weak learners and their corresponding output vectors, requiring classes to share features at each iteration. By training in a cost-sensitive manner, weak learners are invested in separating classes whose discrimination is important, at the expense of less relevant classification boundaries. Additional contributions are a family of loss functions along with proof that our algorithm is Boostable in the theoretical sense, as well as an efficient procedure for growing decision trees for use as weak learners. We evaluate our method on a variety of datasets: a collection of synthetic planar data, common UCI datasets, MNIST digits, SUN scenes, and CUB-200 birds. Results show state-of-the-art performance across all datasets against several strong baselines, including non-boosting multi-class approaches.

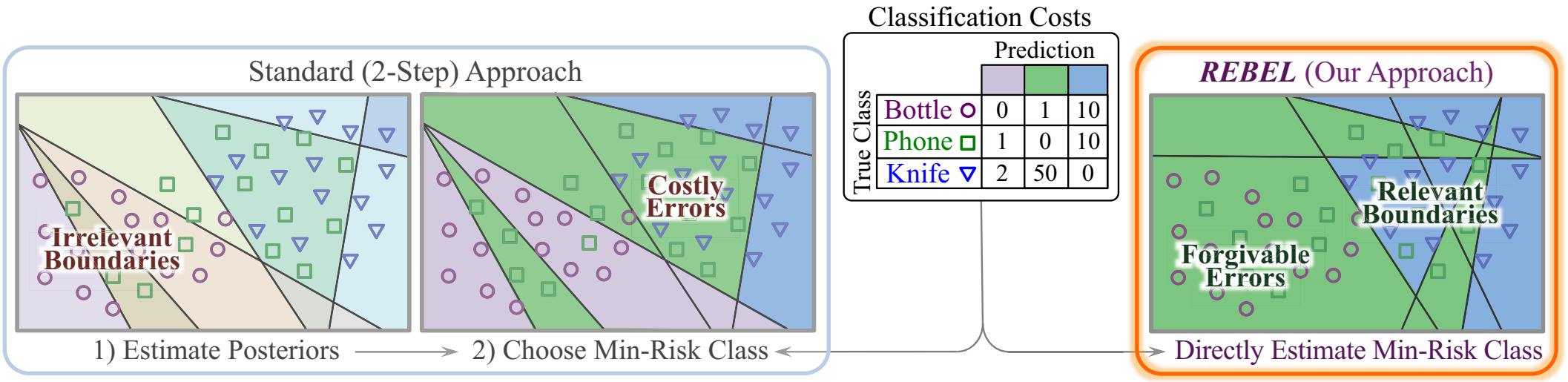

Figure 1: Example scenario of an airport security checkpoint with three classes and corresponding classification costs. With standard training, cost-sensitive classification is a two-step (potentially inefficient) process. Weak learners may be *wasted* on irrelevant (low-cost) boundaries. Using cost-sensitive training (e.g. REBEL), relevant boundaries are set first, giving less weight to the distinction between classes with forgivable errors.

## 1 Introduction

Boosting is one of the most popular off-the-shelf learning techniques in use today. Boosted classifiers are theoretically well understood [34, 20, 21], are easy and quick to train [2], and yield the fastest algorithms when computed as a cascade [6], enabling real-time big-data applications such as object detection and animal tracking [40, 16, 27].

In the case of binary classification, boosting is well understood. From a statistical standpoint, boosting can be seen as iteratively improving an estimator of the underlying posterior distribution of the data [22]. However, the multi-class case is not yet as well understood. Generalizing binary boosting to multi-class often requires various heuristics in order to work (see Sec. 2). Moreover, different errors may incur different costs, with such situations arising in a wide variety of domains such as computer vision, behavior analysis, and medical diagnosis to name a few [12, 9, 32].

An interesting case in point is the categorization of objects naturally organized into a taxonomy, such as birds [7]. Misclassifying taxonomically close categories (eg. confusing two types of duck) is usually more forgivable than distant categories (eg. mistaking a vulture for a duck) [13]. Accordingly, classifiers should be trained to minimize classification cost rather than error rates.

We present a novel approach to multi-class cost-sensitive boosting, offering both a clearer theoretical framework and improved performance over prior art. Instead of first approximating the class-wise posterior distributions and then taking costs into account at test time, our approach directly estimates the class with minimum risk (i.e. minimum expected cost) by using the costs during training. Since our proposed loss function estimates the underlying risk, and in particular, is based on a sum of exponentials, we call our method Risk Estimation Boosting using an Exponential Loss, or *REBEL*.

We prove that REBEL is a true Boosting algorithm in the theoretical sense and outline a method for jointly optimizing decision trees as weak learners, facilitating feature sharing among classes.

In summary, the contributions of our work are:

1. a novel family of loss functions that facilitate multi-class cost-sensitive training (see Sec. 3)
2. a proof that REBEL employs the weak learning condition required for *Boostability* (see Sec. 3)
3. a greedy optimization procedure for growing deeper binary decision trees (see Sec. 4)
4. state-of-the-art results on several datasets for both neutral and cost-sensitive setups (see Sec. 5)

## 2 Related Work

There are two main approaches to multi-class classification: combining multiple binary classifiers, and training multi-class classifiers directly, both of which already exist in the context of boosting.

A multi-class prediction can be achieved by compounding the outputs of multiple binary classifiers. AdaBoost.M2 [21] and AdaBoost.MR [35] train binary classifiers that maximize the margins between pairs of classes. AdaBoost.MH [35] concurrently trains $M$ one-vs-all classifiers, sharing the same weak learners for each binary classifier. Reduction to multiple one-vs-all or one-vs-one classifiers requires augmenting the given datasets, and often results in sub-optimal joint classifiers [33]. ECOC [1] trains multiple binary classifiers and uses error-correcting codes for output classification. Selecting error-codes is often problem dependent and not straightforward; AdaBoost.ERP [26] attempts to find a good trade-off between error-correcting and base learning. JointBoost [39] trains shared binary classifiers, optimizing over all combinations of binary groupings of classes. CD-MCBoost [33] and CW-Boost [36] perform coordinate-descent on a multi-class loss function, thereby focusing each weak learner on a single class. HingeBoost.OC [23] uses output coding combined with a multi-class hinge loss, demonstrating improvements in performance on several datasets. In all of these approaches, the number of classifiers trained (and eventually evaluated) is super-linear in the number of classes, potentially slowing down classification [16].

On the other hand, strong boosted classifiers can be generated using a single chain of multi-class weak learners directly, avoiding the super-linear issue of the aforementioned methods. AdaBoost.M1 [21] is a simple adaptation of binary AdaBoost that makes use of multi-class learners, but has an unreasonably strong (often unachievable) weak-learning condition [29]. SAMME [44] uses a fixed set of codewords to additively generate a score for each class; however, [29] show that it does not satisfy the weak learning conditions required to be *Boostable* (i.e. it is unable to reduce the training error in some situations). GD-MCBoost [33] also uses a fixed set of codewords. AOSO-LogitBoost [38] improves upon previous methods by adaptively separating pairs of classes at each iteration. Struct-Boost [37] combines weak structured learners into a strong structured classifier, applicable to multi-class classification. GLL-Boost [4] exhibits guess-aversion, attempting to avoid outputting equal scores over all classes. However, multi-class weak learners are not as easy to generate and are inherently more complex than their binary counterparts, potentially leading to worse overfitting of the training data. Further, these methods require a sum-to-zero constraint on their output codes, adding to the complexity of computation as well as restricting the classifier's output [28].

As discussed above, many varieties of boosting have been proposed over the years. However, most approaches do not properly deal with the case of cost-sensitive classification. As a workaround, several post-hoc methods exist that interpret classifier output scores as posterior distributions, enabling the estimation of the minimum-risk class as a second step [18, 22, 31]. Approximating posterior distributions is often inaccurate [30], and we claim is unnecessary. Our proposed method is distinguished from prior work, consolidating the following desirable properties into a single framework:

- a novel multi-class loss function that is easily implementable due to its simplicity (see Eq. 2)
- direct estimation of the minimum-risk class via cost-sensitive training; avoiding the need to (inaccurately) approximate a posterior distribution (see Sec. 3)
- a classifier composed of a single chain of binary weak learners that are less complex and share features among all classes, reducing the amount of overfitting [33]
- vector-valued outputs <u>without</u> sum-to-zero constraints simplifying optimization and producing classifiers that can be more expressive
- a novel method for growing binary decision trees as weak learners for use with vector-valued outputs (see Sec. 4), leading to improved performance (see Table 1)

Unifying all the above features into a single framework translates into superior performance as demonstrated in Sec. 5, compared against both prior boosting and non-boosting approaches. Considering that boosting is one of the most widely used supervised classification frameworks, we find that a simple, improved multi-class cost-sensitive classifier is an important contribution to the field.

Although this work focuses on developing a better boosting algorithm; for completeness, we also compare against several other state-of-the-art algorithms: 1-vs-All and multi-class SVMs [10], Structured SVMs [7] and Random Forests [8].

## 3 Approach

In this section, we introduce multi-class classification, formulate an appropriate upper-bounding function, and generalize to deal with cost-sensitive classification. We first define our notation.

| **Notation** | |
|---|---|
| scalars (regular), vectors (bold), constant vectors: | $x$, $\mathbf{x} \equiv [x_1, x_2, ...]$, $\mathbf{0} \equiv [0,0,...]$, $\mathbf{1} \equiv [1,1,...]$ |
| indicator vector, logical indicator function: | $\boldsymbol{\delta}_k$ ($\mathbf{0}$ with a 1 in the $k^{\text{th}}$ entry), $\mathbb{1}(\text{LOGICAL EXPRESSION}) \in \{0,1\}$ |
| inner product, element-wise multiplication: | $\langle \mathbf{x}, \mathbf{v} \rangle$, $\mathbf{x} \odot \mathbf{v}$ |
| element-wise function: | $\mathbf{F}[\mathbf{x}] \equiv [F(x_1), F(x_2), ...]$ |

The goal of multi-class classification is to obtain a function $h$ that correctly predicts the class of queried data points. A data point $\mathbf{x}$ is represented as a feature vector and is associated with a class label $y$. If there are a total of $d$ features and $K$ possible output classes, then:

$$\mathbf{x} \in \mathcal{X} \subseteq \mathbb{R}^d, \quad y \in \mathcal{Y} \equiv \{1, 2, ..., K\}, \quad h : \mathcal{X} \to \mathcal{Y}$$

For a specific $h$, the expected misclassification error is: $\varepsilon \equiv \mathbb{E}\{\mathbb{1}(h(\mathbf{x}) \neq y)\} \equiv \mathbb{E}\{\langle \mathbf{1}-\boldsymbol{\delta}_y, \boldsymbol{\delta}_{h(\mathbf{x})} \rangle\}$

Since $h$ outputs a discrete label, we construct a vector-valued *score function* $\mathbf{H} : \mathcal{X} \to \mathbb{R}^K$, where the index of the maximum value is assigned as the output class: $h(\mathbf{x}) \equiv \arg\max_k \{\langle \mathbf{H}(\mathbf{x}), \boldsymbol{\delta}_k \rangle\}$

In practice, we do not know the underlying distribution of the data; hence, we empirically approximate the expected misclassification error. Given a stochastically sampled training set of $N$ points, the error is approximated as: $\varepsilon \approx \frac{1}{N} \sum_{n=1}^{N} \langle \mathbf{1}-\boldsymbol{\delta}_{y_n}, \boldsymbol{\delta}_{h(\mathbf{x}_n)} \rangle$

Hence, we formulate our problem as: $\mathbf{H}^* = \arg \min_{\mathbf{H} \in \mathcal{H}} \{\varepsilon\}$, where $\mathcal{H}$ is a hypothesis space of possible functions. In this form, directly minimizing the error is infeasible as it is both discontinuous and non-convex [28]. Instead, we propose a family of surrogate loss functions that can tractably be minimized and reach the same optimum as the original error.

Consider any scalar function $F$ that is convex, smooth, upper-bounds the unit step function, and attains the same minimum value (i.e. $F(x) \equiv e^x$, $\log_2(1+e^x)$, $(x+1)^2$, etc.), then the following coupled sum also possesses the above-mentioned properties (see supplement for proof):

$$\langle \mathbf{1}-\boldsymbol{\delta}_y, \boldsymbol{\delta}_{h(\mathbf{x})} \rangle \leq \frac{1}{2} \big( \langle \mathbf{1}-\boldsymbol{\delta}_y, \mathbf{F}[\mathbf{H}(\mathbf{x})] \rangle + \langle \boldsymbol{\delta}_y, \mathbf{F}[-\mathbf{H}(\mathbf{x})] \rangle \big) \qquad (1)$$

As previously discussed, different errors may incur different costs (i.e. letting a knife through a security checkpoint versus a water bottle). Costs can be encoded as a matrix $\underline{\mathbf{C}}$, where each entry $c_{yk} \geq 0$ is the cost of classifying a sample as class $k$ when its true class is $y$. We implicitly assume that correct classification incurs no cost; $c_{yy} \equiv 0$. A cost vector $\mathbf{c}_y$ is defined as the $y^{\text{th}}$ row of the cost matrix. Note that $\mathbf{c}_y$ can be uniquely decomposed as follows (see supplement for details):

$$\mathbf{c}_y = \beta_y \, \mathbf{1} + \sum_{k=1}^{K} b_{ky} \, [\mathbf{1} - \boldsymbol{\delta}_k] \qquad \text{where: } b_{ky} \geq 0, \; b_{jy} = 0 \;\; (\text{for some } j)$$

Accordingly, the empirical risk (i.e. expected cost) can be expanded as:

$$\mathcal{R} \equiv \mathbb{E}\big\{\langle \mathbf{c}_y, \boldsymbol{\delta}_{h(\mathbf{x})} \rangle\big\} \quad \approx \quad \frac{1}{N}\sum_{n=1}^{N} \langle \mathbf{c}_{y_n}, \boldsymbol{\delta}_{h(\mathbf{x}_n)} \rangle \quad \equiv \quad \frac{1}{N}\sum_{n=1}^{N} \beta_{y_n} + \frac{1}{N}\sum_{n=1}^{N}\sum_{k=1}^{K} b_{ky_n} \langle \mathbf{1} - \boldsymbol{\delta}_k, \boldsymbol{\delta}_{h(\mathbf{x}_n)} \rangle$$

Using the bound in (Eq. 1), the risk $\mathcal{R}$ is replaced with a convex upper-bounding surrogate loss:

$$\boxed{\mathcal{L} \equiv \underbrace{\frac{1}{N}\sum_{n=1}^{N}\Big(\beta_{y_n} + \frac{c_n^*}{2}\Big)}_{\mathcal{L}^*} + \frac{1}{2N}\sum_{n=1}^{N}\big(\langle \mathbf{c}_n^+, \mathbf{F}[\mathbf{H}(\mathbf{x}_n)] \rangle + \langle \mathbf{c}_n^-, \mathbf{F}[-\mathbf{H}(\mathbf{x}_n)] \rangle - c_n^*\big)} \tag{2}$$

$$\text{where: } c_n^* \equiv \min_{\mathbf{H}} \{\langle \mathbf{c}_n^+, \mathbf{F}[\mathbf{H}] \rangle + \langle \mathbf{c}_n^-, \mathbf{F}[-\mathbf{H}] \rangle\} \; \text{ and: } \; \mathbf{c}_n^+ \equiv \mathbf{c}_{y_n} - \beta_{y_n} \mathbf{1}, \;\; \mathbf{c}_n^- \equiv (\max\{\mathbf{c}_{y_n}\})\mathbf{1} - \mathbf{c}_{y_n}$$

Up to this point, we have derived a new multi-class cost-sensitive loss function. We now focus on a specific model for $\mathbf{H}$ and outline our corresponding optimization scheme.

### Greedy Iterative Minimization

In this work, we model $\mathbf{H}$ as a weighted sum of binary weak learners. Recall that in standard binary boosting, a strong learner $h_T$ is the cumulative sum of weak learners, where $\alpha_t \in \mathbb{R}$ are scalar weights and $f_t : \mathbb{R}^d \to \{\pm 1\}$ are binary weak learners (chosen from a hypothesis set $\mathcal{F}$), and the final classification is just the sign of $h_T(\mathbf{x})$ [34]. In our model, the scalar weights $\alpha_t$ are extended into $K$-dimensional vectors $\mathbf{a}_t \in \mathbb{R}^K$ as follows:

$$h_T(\mathbf{x}) \equiv \alpha_0 + \sum_{t=1}^{T} f_t(\mathbf{x}) \, \alpha_t \quad \rightarrow \quad \mathbf{H}_T(\mathbf{x}) \equiv \mathbf{a}_0 + \sum_{t=1}^{T} f_t(\mathbf{x}) \, \mathbf{a}_t \tag{3}$$

and the final classifier outputs the index of the maximum-valued element in $\mathbf{H}$. Hence, on the $T{+}1^{\text{th}}$ iteration, we have: $\mathbf{H}_{T+1}(\mathbf{x}) = \mathbf{H}_T(\mathbf{x}) + f_{T+1}(\mathbf{x}) \, \mathbf{a}_{T+1}$. Greedy iterative optimization may be carried out by fixing all parameters from the previous $T$ iterations and minimizing the loss function with respect to the $T{+}1^{\text{th}}$ iteration parameters. From hereon in, we explicitly set our convex upper-bound as: $F(x) \equiv e^x$ due to its simplicity and closed-form solution.

$$\mathcal{L}_f(\mathbf{a}) = \mathcal{L}^* + \frac{1}{2N}\sum_{n=1}^{N}\big(\langle \mathbf{c}_n^+, \mathbf{exp}[\mathbf{H}_T(\mathbf{x}_n) + f(\mathbf{x}_n) \, \mathbf{a}] \rangle + \langle \mathbf{c}_n^-, \mathbf{exp}[-[\mathbf{H}_T(\mathbf{x}_n) + f(\mathbf{x}_n) \, \mathbf{a}]] \rangle - c_n^*\big)$$
$$= \mathcal{L}^* + \frac{1}{2N}\sum_{n=1}^{N}\big(\langle \mathbf{w}_n^+, \mathbf{exp}[f(\mathbf{x}_n) \, \mathbf{a}] \rangle + \langle \mathbf{w}_n^-, \mathbf{exp}[-f(\mathbf{x}_n) \, \mathbf{a}] \rangle - c_n^*\big)$$

with *multi-class weights* $\mathbf{w}^{\pm}$ defined as: $\mathbf{w}_n^+ \equiv \mathbf{c}_n^+ \odot \mathbf{exp}[\mathbf{H}_T(\mathbf{x}_n)]$, $\mathbf{w}_n^- \equiv \mathbf{c}_n^- \odot \mathbf{exp}[-\mathbf{H}_T(\mathbf{x}_n)]$

Joint optimization of $f$ and $\mathbf{a}$ is accomplished by looping over candidate weak learners $f \in \mathcal{F}$ (the hypothesis space of possible weak learners), optimizing the corresponding vector $\mathbf{a}$, and selecting the joint $\{f, \mathbf{a}^*\}$ that minimizes $\mathcal{L}_f(\mathbf{a}^*)$. Note that every element of $\mathbf{a}^*$ is dynamically optimized at every iteration. It is neither a fixed code-word, nor has a sum-to-zero constraint, enhancing the expressiveness of the classifier, ultimately leading to better accuracy (see Sec. 5). Further, in the binary (2-class) case, this framework exactly reduces to binary AdaBoost, LogitBoost, etc. (depending on the choice of $F(x)$; see supplement for details).

### Weak Learning Condition

Although REBEL trains a multi-class classifier, each of its weak learners is a binary classifier. For an algorithm to be *Boostable*, given adequate weak learners (i.e. having at least $50\% + \gamma$ accuracy, where $\gamma > 0$), it must drive the training error to zero [29]. REBEL's weak learning condition is the following (please see supplement for more details):

$$\exists\, \gamma > 0 \quad \text{such that: } \forall\, \mathbf{w}_n^+, \mathbf{w}_n^- \geq \mathbf{0}, \quad \exists\, f \quad \text{satisfying:}$$

$$\left\langle \left| \sum_{n=1}^{N} \left[\mathbf{w}_n^+ - \mathbf{w}_n^-\right] f(\mathbf{x}_n) \right|, \mathbf{1} \right\rangle \geq \gamma \sum_{n=1}^{N} \langle |\mathbf{w}_n^+ - \mathbf{w}_n^-|, \mathbf{1} \rangle$$

For any positive multi-class weights $\mathbf{w}_n^+$ and $\mathbf{w}_n^-$, if there exists a weak learner $f$ such that the above inequality holds, REBEL is guaranteed to minimize the training loss exponentially quickly. REBEL's weak learning condition, although seemingly complex, is equivalent to the binary weak learning condition (please see supplement for proof), implying that REBEL is a true Boosting algorithm in the theoretical sense. The intuition behind this equivalence is that at each iteration, REBEL is actually only making a binary split between two sets of classes. These sets are dynamically chosen at each iteration to lead to optimal separation given the available pool of weak learners.

## 4 Decision Trees

In section 3, we introduced our loss function and an optimization procedure given a hypothesis space of binary weak learners. In this section, we describe our approach specifically for growing useful binary decision trees. Decision trees are simple white-box models, and in particular, shallow trees (i.e. depth $\leq 4$) generalize well and have proven robust in practice [17].

A decision stump $f : \mathcal{X} \to \{\pm 1\}$ can be written as: $f(\mathbf{x}) \equiv \rho\, \text{sign}(\langle \mathbf{x}, \delta_j \rangle - \tau)$ where $\rho \in \{\pm 1\}$ is a polarity, $\delta_j$ selects the $j^{\text{th}}$ feature in $\mathbf{x}$ (out of $d$ possible features), and $\tau \in \mathbb{R}$ is a threshold. Let $\mathcal{H}_1$ denote the hypothesis space of all unique decision stumps on the training data. The input space $\mathcal{X} \subseteq \mathbb{R}^d$ has $d$ dimensions, at times on the order of several thousands (see Sec. 5). Each dimension has a finite number of possible thresholds (at most equaling the number of training samples $N$). In practice, we only consider a fixed number of thresholds $N_\tau$[1]. Finally, there are two possible polarities; hence, $d \times N_\tau \times 2$ unique stumps in $\mathcal{H}_1$. Let $\mathcal{H}_D$ denote the space of unique depth-$D$ trees. For each split in a depth-$D$ tree, both child nodes are depth-($D$–1) trees. This leads to an exponential number of possible trees, totaling $(d \times N_\tau \times 2)^D$ unique trees in $\mathcal{H}_D$. Even on a GigaHertz computer, looping through each possible tree in $\mathcal{H}_D$ is only feasible for $D = 1$. To overcome this issue, we developed a greedy algorithm for using deeper trees as weak learners in our estimator.

### Greedily Growing Trees

By greedily adding one depth-layer at a time, we can grow trees to any depth in a computationally tractable manner. Using our proposed approach, we guarantee that with each added depth-layer, the tree leads to a more accurate strong classifier. Let us assume that we have already jointly trained a depth-$D$ tree $f^{(D)}$ and corresponding output vector $\mathbf{a}^{(D)}$; we now describe how to add the $D{+}1^{\text{th}}$ layer.

1. Replace each leaf node in $f^{(D)}$ with a stump having identical parameters as its parent node, thereby deepening the tree without changing its functionality.
2. Holding $\mathbf{a}^{(D)}$ fixed, optimize each of the $2^D$ newly added stumps. This operation is $O(2^D \times d \times N_\tau)$, resulting in a new tree $f^{(D+1)}$. In the worst case, all added stumps (and hence, overall accuracy) remain unchanged.
3. Fixing $f^{(D+1)}$, minimize $\mathcal{L}(f, \mathbf{a})$ with respect to $\mathbf{a}$, storing the optimal vector as $\mathbf{a}^{(D+1)}$.

Figure 2 illustrates our tree-growing technique, reducing the computational complexity from $10^{6D}$ to $2^D \times d \times N_\tau$, making our classifier fit for use with shallow binary decision trees as weak learners.

[1] We find that $N_\tau \approx 200$ (evenly spaced thresholds over the range of the training data) is sufficient to achieve maximal accuracy in practice. Hence: $d \times N_\tau \times 2 \approx 10^6$ unique stumps in $\mathcal{H}_1$.

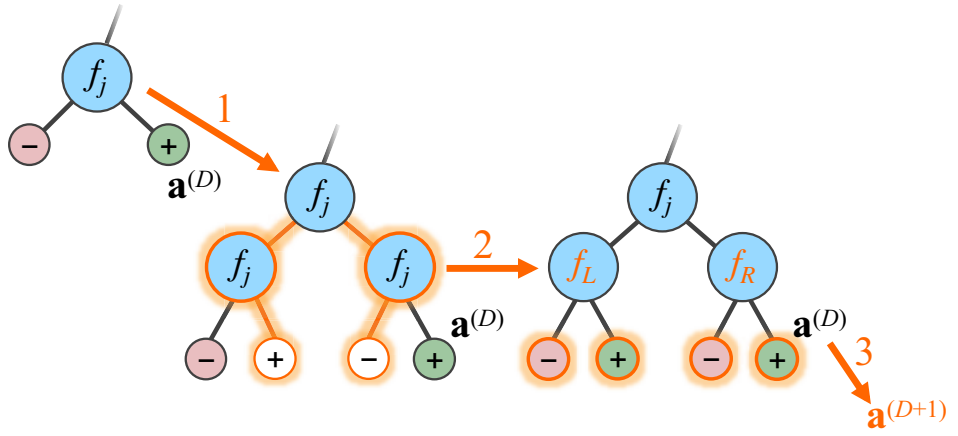

Figure 2: Greedily growing a binary decision tree by one layer. Please see text above for details.

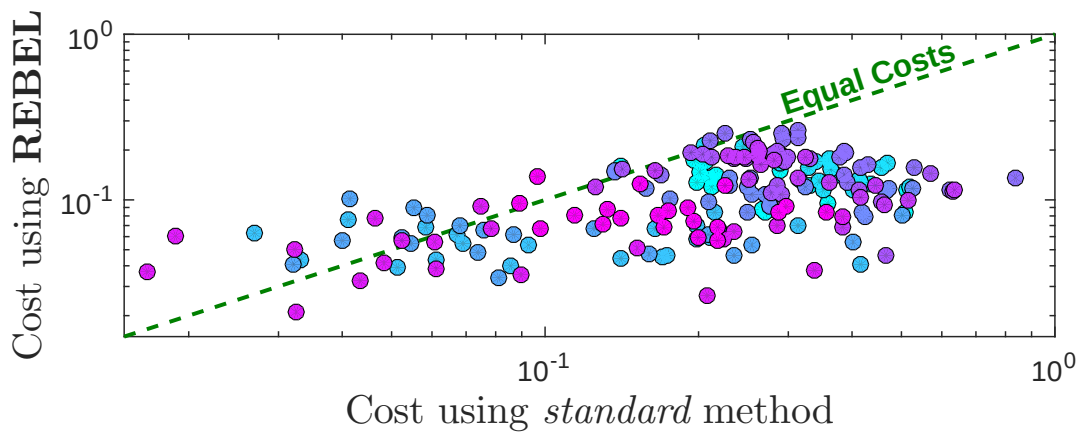

Figure 3: REBEL compared against the standard (two-step) approach. Please see text below for details.

## 5 Experiments

We benchmark our approach on both synthetic and real data, exploring its performance vis-a-vis other methods and demonstrating its functionality. Experiments using synthetic data test REBEL's main properties; please see supplement for corresponding results. Fig. 3 compares REBEL's performance (using cost-sensitive training) to the standard method (i.e. training on just the data, estimating the posterior distributions, and using the costs only at test-time). 10 random datasets and 20 random cost matrices are generated for a total of 200 trials. Each dataset has 1000 training and 500 test points drawn from a mixture of multiple Gaussian clusters (datasets are plotted in the supplement). All cost matrices have zeros along the diagonals and (positive) normally-distributed off-diagonal entries, normalized such that random classification results in unit cost. For each trial, classifiers are trained using 100 stumps. REBEL outperforms the standard method in $\sim$ 90% of the trials, especially on the harder classification problems (i.e. when both methods incur relatively high costs).

### 5.1 Standard multi-class classification on real data

We benchmark REBEL on several UCI datasets [3] and the MNIST handwritten digits [25], using a uniform (cost-neutral) metric. We compared our method against several competitive boosting methods: 1-vs-All AdaBoost and AdaBoost.MH [35], AdaBoost.ECC [15] Struct-Boost [37], CW-Boost [36], and A0S0-LogitBoost [38]. Based on the experimental setup in [37], we use stumps as weak learners; each method having a maximum of 200 weak learners. Five versions of each dataset are generated, randomly split with 50% of the samples for training, 25% for validation (i.e. for finding the optimal number of weak learners), and the remaining 25% for testing. All misclassifications are equally costly; hence, we report percent error with the resulting means and standard deviations in Fig.4. REBEL is the best performing method on three of these five datasets, and is second best to CW-Boost on the two remaining datasets. Even in these commonly-used and *saturated* UCI datasets, we are able to show consistent accuracy compared to previous approaches.

### 5.2 Cost-sensitive classification

As discussed in the introduction, some scenarios require cost-sensitive classification. Among many existing datasets, we chose two current datasets that fall into this category: (1) SUN scene recognition [42] and (2) CUB200 fine-grained bird categorization [41]. Both of these datasets are organized into hierarchies; hence, an apt evaluation metric penalizes each misclassification according to its distance from the true class along the hierarchy.

We compare our method against state-of-the-art boosting methods as well as non-boosting baselines: 1vsAll AdaBoost and AdaBoost.MH [35], AdaBoost.ECC [1], Struct-Boost [37], CW-Boost [36], A0S0-LogitBoost [38], multi-class and 1vsAll SVM [10], Struct-SVM [7], and Random Forests [8].

For Struct-Boost [37], we report performance as published by their authors. For the remaining methods, we downloaded or re-implemented code, cross-validating using the training data to establish optimal parameters; specifically, the number of boosted weak learners (up to 200 on SUN and 2500 on CUB) and tree depth (to a maximum of 4). For the SVMs, we performed grid search through $C$ and $\gamma$ as recommended in [10]. Finally, for the random forest, we performed grid search for tree depth (to a maximum of 7) and number of trees (between 1 and 1000) with combinations not exceeding the number of feature splits allotted to the boosted methods to ensure a fair comparison. For completeness, we report the performance of each method on both the hierarchical misclassification cost as well as uniform cost metric.

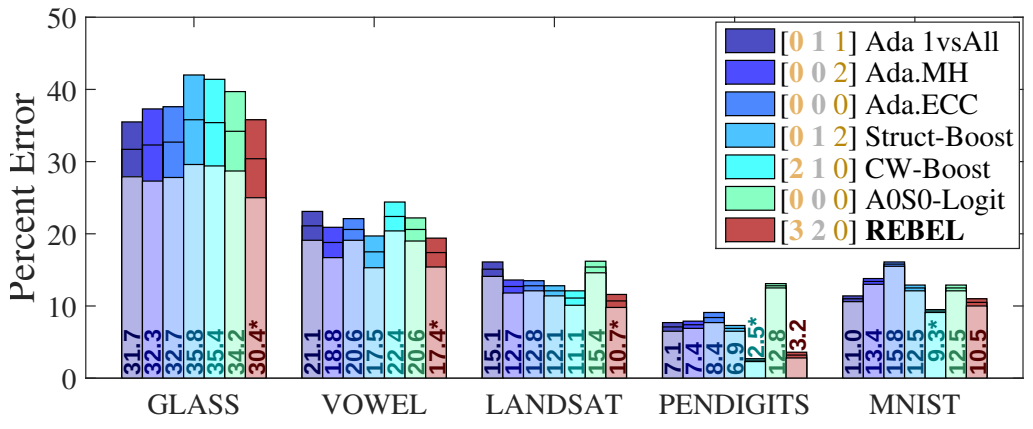

Figure 4: Popular boosting algorithms tested on UCI datasets. Caps on the bars indicate $\pm 1$ std. from the mean. The number of times each method places $1^{\text{st}}$, $2^{\text{nd}}$, and $3^{\text{rd}}$ is indicated in the legend. REBEL consistently places in the top two on every dataset.

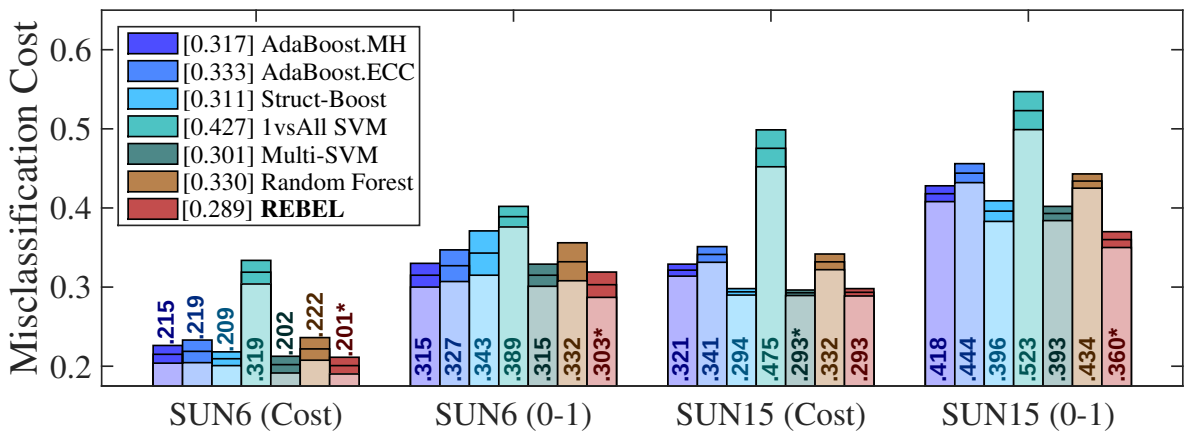

Figure 5: SUN scene recognition datasets, with state-of-the-art classifiers evaluated on a six-class (SUN6) and a fifteen-class subset (SUN15). Each method was trained and evaluated in both a (Cost) cost-sensitive and (0−1) cost-neutral way. REBEL is a consistent top-performing method across all of these subsets.

### SUN scene recognition

We benchmark REBEL on the same two subsets of the SUN database as used in [37], containing six and fifteen classes respectively, and using HOG features as input [11]. For each subset, five random splits were generated with 75% of the samples used for training and 25% for testing. In each of the SUN subsets, the corresponding cost matrices were normalized by the appropriate factor to ensure that random guessing incurred the same cost ($\sim$ 83% in the 6-class subset and $\sim$ 93% in the 15-class subset). The mean and standard deviations over the five splits are shown in Fig. 5.

In cross-validating for optimal parameters on the SUN subsets, REBEL performed best using depth-2 trees as weak learners, see Table 1. REBEL is consistently the top-performing method on all of the subsets, edging out multi-class SVM and the other boosting methods. Other than the one-vs-all SVM which consistently under-performs, most of the methods lead to similar accuracies.

| | Stumps | Depth-2 | Depth-3 | Depth-4 |
|---|---|---|---|---|
| SUN6(0–1) | 32.8% | **30.3%** | 34.2% | 35.0% |
| SUN6(cost) | 0.363 | **0.323** | 0.327 | 0.325 |

| | Stumps | Depth-2 | Depth-3 | Depth-4 |
|---|---|---|---|---|
| CUB(0–1) | 21.4% | **20.9%** | 22.1% | 22.9% |
| CUB(cost) | 0.291 | 0.291 | 0.286 | **0.26** |

Table 1: REBEL misclassification errors on SUN-6 and CUB datasets using different tree depths. Deeper trees consistently outperform stumps, giving credit to our tree-growing method introduced in Sec. 4.

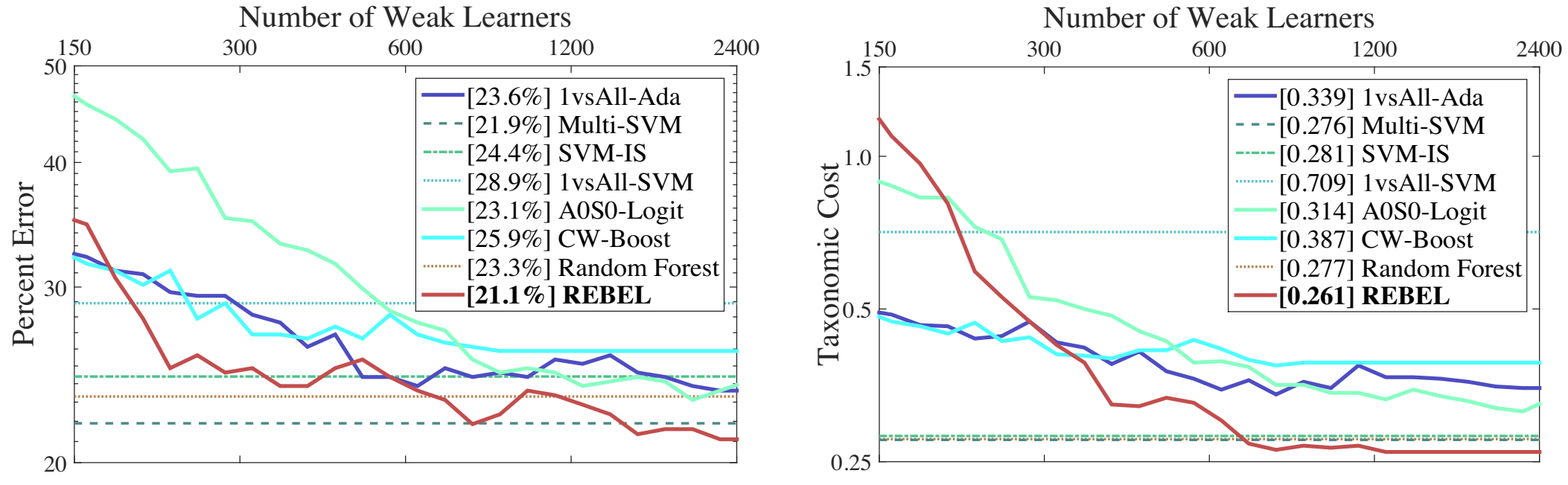

(a) Misclassification error (0–1)

(b) Cost-sensitive (Cost)

Figure 6: CUB200 fine-grained bird categorization results. Several state-of-the-art classifiers evaluated on a standard fourteen-class subset. All methods are trained and evaluated (a) cost-neutrally and (b) cost-sensitively, tree depths for each boosting method are chosen by cross-validation. The lowest cost of each method on the test set is indicated in square brackets. REBEL with decision trees is the best performing method in both cases.

### CUB fine-grained bird categorization

We further benchmark REBEL on the CUB200 fine-grained bird categorization dataset, using a fourteen-class *Woodpecker* subset as in [19, 5, 43, 14]. Each bird in the subset is represented as a 4096-feature vector; the output of a pre-trained Convolutional Neural Network [24]. The subset consists of 392 training and 398 test samples.

Classification errors are shown in Fig. 6. Boosting methods are plotted as a function of the number of weak learners and non-boosting methods are plotted as horizontal lines. REBEL is the best performing method in both cases. All other boosting methods saturated between 20% and 40% higher costs than REBEL; significantly worse. REBEL also outperformed multi-class SVM, albeit by much smaller margins. Since this subset of birds has a taxonomy that is particularly shallow, the taxonomic cost matrix is not that different from that of uniform errors; hence, we do not expect drastic improvements in performance and are satisfied with these *modest* gains.

In cross-validating for optimal parameters on CUB, REBEL performed best using depth-2 trees as weak learners for the cost-neutral experiment and depth-4 trees for the cost-sensitive experiment (see Table 1), validating the benefit of our decision tree growing approach introduced in Sec.4.

# 6 Discussion

Our experimental results indicate that REBEL is as good as – and even outperforms – other state-of-the-art classifiers. But (1) why does it work so well? And (2) why does it also outperform existing classifiers in cost-insensitive situations (in which REBEL seemingly has no *advantage*)?

(1) We are pleased with REBEL's performance and seek a better understanding. Our current explanation relates to the form of our model and agrees with empirical results; summarized as follows:

(a) In each iteration, REBEL determines a single binary split $f_t$ (and the corresponding optimal vector $\mathbf{a}_t$). This forces feature sharing across classes. Experiments show superior performance against baselines that do not require feature sharing (see Figs. 4,5: REBEL vs 1vAll, ECC, CW). Feature sharing is a form of regularization [39, 33].

(b) Avoiding sum-to-zero constraints on the associated vector $\mathbf{a}_t$ is empirically advantageous. The additional degree of freedom lets each weak learner and vector pair be more expressive, leading to a greater reduction in loss at each iteration (see Figs. 4,5: REBEL vs AOSO, Fig. 6: REBEL's quick initial descent in error/cost).

Therefore (a) and (b) act in opposite directions and the balance of the two leads to beneficial results. This finding is also theoretically justified since REBEL employs the Binary Weak Learning Condition in [29], thereby neither overfitting nor underfitting.

(2) The proposed surrogate loss function (Eq. 2) is set up to incorporate classification costs during training. However, misclassification cost is equivalent to classification error when the cost matrix has uniform costs. In this case, the minimum-risk class is the max-posterior class. Hence, there is no reason to expect better or worse performance than other methods that minimize classification error. The fact that our our method performs slightly better on the datasets we tested (see Fig. 4) is a further indication that our balance of regularization and expressiveness (point (1) above) is advantageous.

# 7 Conclusions

We presented a multi-class cost-sensitive boosting framework with a novel family of simple surrogate loss functions. This framework directly models the minimum-risk class without explicitly approximating a posterior distribution. Training is based on minimizing classification costs, as specified by the user (e.g. following taxonomic distance). Using an exponential-based loss function, we derived and implemented REBEL.

REBEL unifies the best qualities from a number of algorithms. It is conceptually simple, and optimizes feature sharing among classes. We found that REBEL is able to focus on important distinctions (those with costly errors) in exchange for more forgivable ones. We prove that REBEL employs the weak learning condition required to be a true Boosting algorithm in the theoretical sense.

We compared REBEL to several state-of-the-art boosting and non-boosting methods, showing improvements on the various datasets used. Several images from the SUN and CUB-200 datasets with highest and lowest corresponding risk-scores (i.e. $\mathbf{H}(\mathbf{x})$) are plotted in the supplement. Being based on boosting, our method may be implemented as a cascade to yield very fast classification, useful in real-time applications. Our technique therefore holds the promise of producing classifiers that are as accurate and faster than the state-of-the-art.

# Supplementary Materials

| **Notation** | |
|---|---|
| scalars (regular), vectors (bold), constant vectors: | $x,\;\; \mathbf{x} \equiv [x_1, x_2, ...],\;\; \mathbf{0} \equiv [0,0, ...],\;\; \mathbf{1} \equiv [1,1, ...]$ |
| indicator vector, logical indicator function: | $\boldsymbol{\delta}_k$ ($\mathbf{0}$ with a 1 in the $k^{\text{th}}$ entry), $\mathbb{1}(\text{LOGICAL EXPRESSION}) \in \{0,1\}$ |
| inner product, element-wise multiplication: | $\langle \mathbf{x}, \mathbf{v} \rangle,\;\; \mathbf{x} \odot \mathbf{v}$ |
| element-wise function: | $\mathbf{F}[\mathbf{x}] \equiv [F(x_1), F(x_2), ...]$ |

## S.1 Coupled Sum

Let the class of functions $F : \mathbb{R} \to \mathbb{R}$ be convex, upper-bounding of the unit step function, and having the same optimal value of 0:

$$\frac{\partial^2 F}{\partial x^2} \geq 0, \quad F(x) \geq \mathbb{1}(x \geq 0), \quad \inf_x \{F(x)\} = 0$$

Define $\sigma$ as the following coupled sum:

$$\sigma(\mathbf{H}; y) \equiv \frac{\langle \mathbf{1} - \boldsymbol{\delta}_y, \mathbf{F}[\mathbf{H}] \rangle + \langle \boldsymbol{\delta}_y, \mathbf{F}[-\mathbf{H}] \rangle}{2}$$

Then $\sigma$ is: (1) convex, (2) tight at the optimum, and (3) upper-bounding.

**Claim (1)**: $\sigma(\mathbf{H}; y)$ is convex in $\mathbf{H}$.

**Proof**:

$$\sigma(\mathbf{H}; y) \equiv \frac{1}{2}\Big(F(-H_y) + \sum_{k \neq y} F(H_k)\Big)$$

half of the sum of convex functions is also convex.

**Claim (2)**: $\inf_{\mathbf{H}} \big\{\sigma(\mathbf{H}; y)\big\} = 0$

**Proof**:

$$\lim_{\theta \to \infty} \sigma(\theta\, [\mathbf{1} - 2\, \boldsymbol{\delta}_y]; y) = 0$$

**Claim (3)**: $\sigma\big(\mathbf{H}(\mathbf{x}); y\big) \geq \mathbb{1}\big(y \neq \operatorname{argmax}_k \{\langle \mathbf{H}(\mathbf{x}), \boldsymbol{\delta}_k \rangle\}\big)$

**Proof**:

$$\text{Let:} \qquad j = \arg\max_{k \neq y} \Big\{\frac{-H_y + H_k}{2}\Big\}, \quad z = \frac{-H_y + H_j}{2}$$

Then:

$$\sigma(\mathbf{H}; y) \equiv \frac{1}{2}\Big(F(-H_y) + \sum_{k \neq y} F(H_k)\Big) \geq \underbrace{\frac{1}{2}\big(F(-H_y) + F(H_j)\big) \geq F\Big(\frac{-H_y + H_j}{2}\Big)}_{\text{due to Jensen's inequality}} = F(z)$$

**Case (a)**: correct classification *(should upper-bound* 0*)*

$$\forall\, k \neq y,\; H_y > H_k \quad \Rightarrow \quad z < 0 \qquad \Rightarrow \quad \sigma(\mathbf{H}; y) \geq F(z) \geq \mathbb{1}(z \geq 0) = 0$$

**Case (b)**: incorrect classification *(should upper-bound* 1*)*

$$\exists\, j \;\text{ s.t. }\; H_j \geq H_y \quad \Rightarrow \quad z \geq 0 \qquad \Rightarrow \quad \sigma(\mathbf{H}; y) \geq F(z) \geq \mathbb{1}(z \geq 0) = 1$$

## S.2 Exponential Function

Note that in this work, we explicitly set: $F(x) \equiv e^x$

The exponential function is smooth, convex, upper-bounds the unit step function, and attains a minimum value of 0 for: $x^* = -\infty$, hence, it satisfies all required properties and is suitable for use.

## S.3 Cost Vector Decomposition

**Claim**: Any cost vector $\mathbf{c}_y$ can be decomposed as:

$$\mathbf{c}_y = \beta_y \, \mathbf{1} + \sum_{k=1}^{K} b_{ky} \, [\mathbf{1} - \boldsymbol{\delta}_k] \qquad \text{where: } \beta_y, b_{ky} \in \mathbb{R}, \;\; b_{ky} \geq 0$$

This decomposition is unique when $\sum_k b_{ky}$ is minimal.

**Proof**: $\mathbf{c}_y$ is uniquely decomposed as:

$$\mathbf{c}_y = \sum_{k=1}^{K} c_{ky} \, \boldsymbol{\delta}_k = \Big(\sum_{k=1}^{K} c_{ky}\Big) \mathbf{1} - \sum_{k=1}^{K} c_{ky} [\mathbf{1} - \boldsymbol{\delta}_k]$$

An extra degree of freedom $\phi_y$ is introduced by adding and subtracting the term:

$$0 = -(K-1)\phi_y \, \mathbf{1} + \sum_{k=1}^{K} \phi_y \, [\mathbf{1} - \boldsymbol{\delta}_k]$$

Hence:

$$\mathbf{c}_y = \underbrace{\left( \Big(\sum_{k=1}^{K} c_{ky}\Big) - (K-1)\phi_y \right)}_{\beta_y} \mathbf{1} + \sum_{k=1}^{K} \underbrace{(\phi_y - c_{ky})}_{b_{ky}} [\mathbf{1} - \boldsymbol{\delta}_k]$$

Satisfying: $b_{ky} \geq 0$ implies that: $\phi_y \geq \max_k \{c_{ky}\}$

Finally, minimizing the sum of the $b_{ky}$:

$$\sum_{k=1}^{K} b_{ky} = \sum_{k=1}^{K} (\phi_y - c_{ky}) = K\phi_y - \overbrace{\Big(\sum_{k=1}^{K} c_{ky}\Big)}^{\text{constant}}$$

the smaller the $\phi_y$, the smaller the overall sum; hence, for optimality:

$$\phi_y = \max_k \{c_{ky}\} \qquad \Rightarrow \quad \sum_{k=1}^{K} b_{ky} = K\phi_y - \big(\beta_y + (K-1)\phi_y\big) = \max_k \{c_{ky}\} - \beta_y$$

## S.4 Risk Surrogate

The risk (expected cost) is empirically approximated as: $\mathcal{R} \approx \mathcal{R}_{\text{EMP}} \equiv \frac{1}{N} \sum_{n=1}^{N} \langle \mathbf{c}_{y_n}, \boldsymbol{\delta}_{h(\mathbf{x}_n)} \rangle$

Decomposing the risk into components: $\mathcal{R}_{\text{EMP}} \equiv \frac{1}{N} \sum_{n=1}^{N} \Big( \beta_{y_n} + \sum_{k=1}^{K} b_{ky_n} \langle \mathbf{1} - \boldsymbol{\delta}_k, \boldsymbol{\delta}_{h(\mathbf{x}_n)} \rangle \Big)$

Substituting the convex upper-bounding surrogate loss:

$$\begin{aligned} \mathcal{L} &\equiv \frac{1}{N} \sum_{n=1}^{N} \beta_{y_n} + \frac{1}{2N} \sum_{n=1}^{N} \sum_{k=1}^{K} b_{ky_n} \big( \langle \mathbf{1} - \boldsymbol{\delta}_k, \mathbf{F}[\mathbf{H}(\mathbf{x}_n)] \rangle + \langle \boldsymbol{\delta}_k, \mathbf{F}[-\mathbf{H}(\mathbf{x}_n)] \rangle \big) \\ &= \underbrace{\frac{1}{N} \sum_{n=1}^{N} \Big( \beta_{y_n} + \frac{c_n^*}{2} \Big)}_{\mathcal{L}^*} + \frac{1}{2N} \sum_{n=1}^{N} \big( \langle \mathbf{c}_n^+, \mathbf{F}[\mathbf{H}(\mathbf{x}_n)] \rangle + \langle \mathbf{c}_n^-, \mathbf{F}[-\mathbf{H}(\mathbf{x}_n)] \rangle - c_n^* \big) \end{aligned} \tag{4}$$

where: $$c^*_n \equiv \min_{\mathbf{H}}\{\langle \mathbf{c}^+_n, \mathbf{F}[\mathbf{H}]\rangle + \langle \mathbf{c}^-_n, \mathbf{F}[-\mathbf{H}]\rangle\}, \quad \text{(where } \mathbf{H}^*_n \text{ is the argmin)}$$

$$\mathbf{c}^+_n \equiv \sum_{k=1}^{K} b_{ky_n}[\mathbf{1}-\boldsymbol{\delta}_k] = \mathbf{c}_{y_n} - \beta_{y_n}\mathbf{1}$$

$$\begin{aligned}\mathbf{c}^-_n &\equiv \sum_{k=1}^{K} b_{ky_n}\boldsymbol{\delta}_k = \Big(\sum_{k=1}^{K} b_{ky_n}\Big)\mathbf{1} - \mathbf{c}^+_n \\ &= \big(\max_k\{c_{ky_n}\} - \beta_{y_n}\big)\,\mathbf{1} - [\mathbf{c}_{y_n} - \beta_{y_n}\mathbf{1}] \\ &= (\max_k\{c_{ky_n}\})\,\mathbf{1} - \mathbf{c}_{y_n}\end{aligned}$$

Note that using this construction, $\mathbf{c}^+_n,\ \mathbf{c}^-_n \geq \mathbf{0}$, and further, when the score vectors are optimal for every training sample, i.e. $\mathbf{H}(\mathbf{x}_n) = \mathbf{H}^*_n\ \forall n$, then the maximum-scoring class is always correct; hence, classification cost (i.e. training error) is 0.

Even when a score vector is not equal to its optimal value but still has the correct maximum-scoring class, the training cost is still 0. Specifically, if $\langle \mathbf{H}(\mathbf{x}_n), \boldsymbol{\delta}_{y_n}\rangle = \langle \mathbf{H}(\mathbf{x}_n), \boldsymbol{\delta}_k\rangle = h$ even for a single training sample $n$ and class $k \neq y_n$, then the training cost is greater than 0. Consequently, let $\mathcal{L}_\bullet$ be the minimal loss for which this can occur:

$$\mathcal{L}_\bullet \equiv \mathcal{L}^* + \min_{\substack{n,k\neq y_n\\ h,h_y,h_k}}\Big\{\big((c^+_{ny_n}+c^+_{nk})F(h) + (c^-_{ny_n}+c^-_{nk})F(-h)\big) - \big(c^+_{ny_n}F(h_y) + c^-_{ny_n}F(-h_y)\big) - \big(c^+_{nk}F(h_k) + c^-_{nk}F(-h_k)\big)\Big\} > \mathcal{L}^*$$

Hence: $$\mathcal{L} < \mathcal{L}_\bullet \ \Rightarrow\ \mathcal{R}_{\text{EMP}} = 0 \tag{5}$$

## S.5 Exponential Loss

Using the exponential function as our base function $F$, the loss from Eq. 4 can be expressed as:

$$\mathcal{L} = \mathcal{L}^* + \frac{1}{2N}\sum_{n=1}^{N}\big(\langle \mathbf{c}^+_n, \mathbf{exp}[\mathbf{H}(\mathbf{x}_n)]\rangle + \langle \mathbf{c}^-_n, \mathbf{exp}[-\mathbf{H}(\mathbf{x}_n)]\rangle - c^*_n\big)$$

Determining the optimal score $\mathbf{H}^*_n$ for each data point by setting $\nabla_{\mathbf{H}}\mathcal{L} = \mathbf{0}$:

$$\mathbf{c}^+_n \odot \mathbf{exp}[\mathbf{H}^*_n] = \mathbf{c}^-_n \odot \mathbf{exp}[-\mathbf{H}^*_n] \quad\Rightarrow\quad \mathbf{H}^*_n = \frac{1}{2}\big[\mathbf{ln}[\mathbf{c}^-_n] - \mathbf{ln}[\mathbf{c}^+_n]\big] \quad\Rightarrow\quad c^*_n = 2\langle\sqrt{\mathbf{c}^+_n \odot \mathbf{c}^-_n}, \mathbf{1}\rangle$$

Given a cost matrix, we can explicitly compute $\mathcal{L}_\bullet$:

$$\mathcal{L}_\bullet \equiv \mathcal{L}^* + \min_{n,k\neq y_n}\Big\{2\sqrt{(c^+_{ny_n}+c^+_{nk})(c^-_{ny_n}+c^-_{nk})} - 2\sqrt{c^+_{ny_n}c^-_{ny_n}} - 2\sqrt{c^+_{nk}c^-_{nk}}\Big\}$$

## S.6 Greedy Iterative Minimization

After having trained $T$ iterations, given a weak learner $f$ and vector $\mathbf{a}$, the loss is:

$$\begin{aligned}\mathcal{L}_f(\mathbf{a}) &= \mathcal{L}^* + \frac{1}{2N}\sum_{n=1}^{N}\big(\langle \mathbf{c}^+_n, \mathbf{exp}[\mathbf{H}_T(\mathbf{x}_n) + f(\mathbf{x}_n)\,\mathbf{a}]\rangle + \langle \mathbf{c}^-_n, \mathbf{exp}[-[\mathbf{H}_T(\mathbf{x}_n) + f(\mathbf{x}_n)\,\mathbf{a}]]\rangle - c^*_n\big) \\ &= \mathcal{L}^* + \frac{1}{2N}\sum_{n=1}^{N}\big(\langle \mathbf{w}^+_n, \mathbf{exp}[+f(\mathbf{x}_n)\,\mathbf{a}]\rangle + \langle \mathbf{w}^-_n, \mathbf{exp}[-f(\mathbf{x}_n)\,\mathbf{a}]\rangle - c^*_n\big)\end{aligned} \tag{6}$$

where *multi-class weights* $\mathbf{w}^+_n$ and $\mathbf{w}^-_n$ are defined as:

$$\mathbf{w}^+_n \equiv \mathbf{c}^+_n \odot \mathbf{exp}[\mathbf{H}_T(\mathbf{x}_n)], \quad \mathbf{w}^-_n \equiv \mathbf{c}^-_n \odot \mathbf{exp}[-\mathbf{H}_T(\mathbf{x}_n)]$$

To determine the optimal vector $\mathbf{\mathring{a}}^*$, the loss (Eq. 6) can be simplified as:

$$\mathcal{L}_f(\mathbf{a}) = \mathcal{L}^* + \langle \mathbf{s}_f^+, \mathbf{exp}[\mathbf{a}] \rangle + \langle \mathbf{s}_f^-, \mathbf{exp}[-\mathbf{a}] \rangle - c^* \tag{7}$$

$$\text{where:} \quad \mathbf{s}_f^{\pm} \equiv \frac{1}{2N}\sum_{n=1}^{N}\left[\mathbb{1}(f(\mathbf{x}_n)=+1)\mathbf{w}_n^{\pm} + \mathbb{1}(f(\mathbf{x}_n)=-1)\mathbf{w}_n^{\mp}\right], \quad c^* \equiv \frac{1}{2N}\sum_{n=1}^{N} c_n^*$$

hence, solving for optimal $\mathbf{\mathring{a}}^*$:

$$\nabla_{\mathbf{a}}\mathcal{L}_f(\mathbf{\mathring{a}}^*) = \mathbf{0} \quad \Rightarrow \quad \mathbf{s}_f^+ \odot \mathbf{exp}[\mathbf{\mathring{a}}^*] = \mathbf{s}_f^- \odot \mathbf{exp}[-\mathbf{\mathring{a}}^*] \quad \Rightarrow \quad \mathbf{\mathring{a}}^* = \frac{1}{2}\left[\mathbf{ln}[\mathbf{s}_f^-] - \mathbf{ln}[\mathbf{s}_f^+]\right]$$

$$\therefore \quad \mathcal{L}_f(\mathbf{\mathring{a}}^*) = \mathcal{L}^* + 2\langle \sqrt{\mathbf{s}_f^+ \odot \mathbf{s}_f^-}, \mathbf{1} \rangle - c^*$$

### S.6.1 Minimizing REBEL's Loss

Assume that: $\mathcal{L}_f(\mathbf{0}) \geq \mathcal{L}_\bullet$ hence, from Eq. 7: $\langle \mathbf{s}_f^+ + \mathbf{s}_f^-, \mathbf{1} \rangle - c^* \geq \mathcal{L}_\bullet - \mathcal{L}^*$

Then, for some $\gamma > 0$, the condition: $\langle |\mathbf{s}_f^+ - \mathbf{s}_f^-|, \mathbf{1} \rangle \geq \gamma(\langle \mathbf{s}_f^+ + \mathbf{s}_f^-, \mathbf{1} \rangle - c^*)$ implies the following:

$$\langle |\mathbf{s}_f^+ - \mathbf{s}_f^-|, \mathbf{1} \rangle \geq \phi \langle \mathbf{s}_f^+ + \mathbf{s}_f^-, \mathbf{1} \rangle \qquad \text{where: } \phi \equiv \gamma\left(1 - \frac{c^*}{\mathcal{L}_\bullet - \mathcal{L}^* + c^*}\right) > 0$$

$$\langle |\mathbf{s}_f^+ - \mathbf{s}_f^-|, \mathbf{1} \rangle^2 \geq \phi^2 \langle \mathbf{s}_f^+ + \mathbf{s}_f^-, \mathbf{1} \rangle^2$$

$$\Rightarrow \quad (1-\phi^2)\langle \mathbf{s}_f^+ + \mathbf{s}_f^-, \mathbf{1} \rangle^2 \geq \langle \mathbf{s}_f^+ + \mathbf{s}_f^-, \mathbf{1} \rangle^2 - \langle |\mathbf{s}_f^+ - \mathbf{s}_f^-|, \mathbf{1} \rangle^2 \equiv \langle 2\mathbf{s}_f^+, \mathbf{1} \rangle \langle 2\mathbf{s}_f^-, \mathbf{1} \rangle$$

$$\Rightarrow \quad \sqrt{1-\phi^2}\,\langle \mathbf{s}_f^+ + \mathbf{s}_f^-, \mathbf{1} \rangle \geq 2\sqrt{\langle \mathbf{s}_f^+, \mathbf{1} \rangle \langle \mathbf{s}_f^-, \mathbf{1} \rangle}$$

Further, since the square-root function is concave, then by Jensen's inequality:

$$\frac{\sum_{k=1}^{K} s_{fk}^+ \sqrt{\frac{s_{fk}^-}{s_{fk}^+}}}{\sum_{k=1}^{K} s_{fk}^+} \leq \sqrt{\frac{\sum_{k=1}^{K} s_{fk}^+ \left(\frac{s_{fk}^-}{s_{fk}^+}\right)}{\sum_{k=1}^{K} s_{fk}^+}} \quad \Rightarrow \quad \sum_{k=1}^{K} s_{fk}^+ \sqrt{\frac{s_{fk}^-}{s_{fk}^+}} \leq \sqrt{\sum_{k=1}^{K} s_{fk}^+}\sqrt{\sum_{k=1}^{K} s_{fk}^+ \left(\frac{s_{fk}^-}{s_{fk}^+}\right)}$$

$$\therefore \quad \langle \sqrt{\mathbf{s}_f^+ \odot \mathbf{s}_f^-}, \mathbf{1} \rangle \leq \sqrt{\langle \mathbf{s}_f^+, \mathbf{1} \rangle \langle \mathbf{s}_f^-, \mathbf{1} \rangle}$$

Combining these results:

$$\langle |\mathbf{s}_f^+ - \mathbf{s}_f^-|, \mathbf{1} \rangle \geq \gamma(\langle \mathbf{s}_f^+ + \mathbf{s}_f^-, \mathbf{1} \rangle - c^*) \quad \Rightarrow \quad 2\langle \sqrt{\mathbf{s}_f^+ \odot \mathbf{s}_f^-}, \mathbf{1} \rangle \leq \sqrt{1-\phi^2}\,\langle \mathbf{s}_f^+ + \mathbf{s}_f^-, \mathbf{1} \rangle$$

$$\Rightarrow \quad 2\langle \sqrt{\mathbf{s}_f^+ \odot \mathbf{s}_f^-}, \mathbf{1} \rangle - c^* \leq \sqrt{1-\phi^2}\,(\langle \mathbf{s}_f^+ + \mathbf{s}_f^-, \mathbf{1} \rangle - c^*) \tag{8}$$

leading to the following result (while $\mathcal{L}_T \geq \mathcal{L}_\bullet$):

$$\frac{\mathcal{L}_f(\mathbf{\mathring{a}}^*) - \mathcal{L}^*}{\mathcal{L}_f(\mathbf{0}) - \mathcal{L}^*} \leq \sqrt{1-\phi^2} \leq e^{-\frac{\phi^2}{2}} \quad \Rightarrow \quad \frac{\mathcal{L}_T - \mathcal{L}^*}{\mathcal{L}_0 - \mathcal{L}^*} \leq \left(e^{-\frac{\phi^2}{2}}\right)^T$$

hence, REBEL minimizes the training error at an exponential rate in the number of iterations $T$.

## S.7 Weak Learning Conditions

When an algorithm is *Boostable*, it is able to minimize the training error exponentially fast without imposing too-weak or too-strong requirements on the available weak learners [29] .

For binary classification, the weak learning condition requires that there always exist a weak learner $f : \mathcal{X} \to \{\pm 1\}$ that incurs less cost than random guessing. Given a training set $\{\mathbf{x}_n, y_n\}_{n=1}^N$, then for the $n^{\text{th}}$ sample, let $c_n$ and $\bar{c}_n$ denote the costs of correct and incorrect classification respectively.

Let $\gamma$ denote the *edge*; how much better than random a weak learner must perform. Then, for an algorithm to be Boostable, it must employ the following Weak Learning Condition (WLC):

$$\exists\, \gamma > 0 \quad \text{such that:} \quad \forall\, c_n \leq \overline{c}_n, \;\; \exists\, f \;\; \text{satisfying:}$$
$$\sum_{n=1}^{N} \left( \Big( \frac{1+y_n f(\mathbf{x}_n)}{2} \Big) c_n + \Big( \frac{1-y_n f(\mathbf{x}_n)}{2} \Big) \overline{c}_n \right) \leq \sum_{n=1}^{N} \left( \Big( \frac{1+\gamma}{2} \Big) c_n + \Big( \frac{1-\gamma}{2} \Big) \overline{c}_n \right)$$

by substituting $w_n \equiv \overline{c}_n - c_n$, this is equivalent to:

$$\exists\, \gamma > 0 \quad \text{such that:} \quad \forall\, w_n \geq 0, \;\; \exists\, f \;\; \text{satisfying:} \qquad \sum_{n=1}^{N} w_n y_n f(\mathbf{x}_n) \geq \gamma \sum_{n=1}^{N} w_n \tag{9}$$

### S.7.1 REBEL's Weak Learning Condition

The condition from Eq. 8 above can be expressed as:

$$\Big\langle \Big| \sum_{n=1}^{N} [\mathbf{w}_n^+ - \mathbf{w}_n^-] f(\mathbf{x}_n) \Big|, \mathbf{1} \Big\rangle \geq \gamma \sum_{n=1}^{N} (\langle \mathbf{w}_n^+ + \mathbf{w}_n^-, \mathbf{1} \rangle - c_n^*) \tag{10}$$

$$\text{by expanding:} \quad \mathbf{s}_f^+ - \mathbf{s}_f^- \equiv \sum_{n=1}^{N} [\mathbf{w}_n^+ - \mathbf{w}_n^-] f(\mathbf{x}_n), \quad \text{and:} \quad \mathbf{s}_f^+ + \mathbf{s}_f^- \equiv \sum_{n=1}^{N} [\mathbf{w}_n^+ + \mathbf{w}_n^-]$$

Further, note that:

$$|\mathbf{w}_n^- - \mathbf{w}_n^+| = \Big[ \sqrt{\mathbf{w}_n^-} + \sqrt{\mathbf{w}_n^+} \Big] \Big| \sqrt{\mathbf{w}_n^-} - \sqrt{\mathbf{w}_n^+} \Big| \geq \Big| \sqrt{\mathbf{w}_n^-} - \sqrt{\mathbf{w}_n^+} \Big|^2 \equiv [\mathbf{w}_n^- + \mathbf{w}_n^+] - 2\sqrt{\mathbf{w}_n^+ \odot \mathbf{w}_n^-}$$
$$\therefore \quad \sum_{n=1}^{N} \langle |\mathbf{w}_n^- - \mathbf{w}_n^+|, \mathbf{1} \rangle \geq \sum_{n=1}^{N} \left( \langle \mathbf{w}_n^+ + \mathbf{w}_n^-, \mathbf{1} \rangle - c_n^* \right)$$

Consequently, REBEL's weak learning condition is:

$$\exists\, \gamma > 0 \quad \text{such that:} \quad \forall\, \mathbf{w}_n^+, \mathbf{w}_n^- \geq \mathbf{0}, \quad \exists\, f \quad \text{satisfying:}$$
$$\Big\langle \Big| \sum_{n=1}^{N} [\mathbf{w}_n^+ - \mathbf{w}_n^-] f(\mathbf{x}_n) \Big|, \mathbf{1} \Big\rangle \geq \gamma \sum_{n=1}^{N} \langle |\mathbf{w}_n^+ - \mathbf{w}_n^-|, \mathbf{1} \rangle \tag{11}$$

Therefore, if the loss $\mathcal{L} < \mathcal{L}_\bullet$, then the training cost is guaranteed to already be minimal, otherwise (i.e. $\mathcal{L} \geq \mathcal{L}_\bullet$), the training cost is guaranteed to decrease at an exponential rate with each iteration.

### S.7.2 Equivalence of Binary WLC and REBEL's WLC

To prove that the two weak learning conditions are equivalent, we (a) assume that REBEL's WLC holds and show that it implies the Binary WLC, and (b) assume that the Binary WLC holds and show that it implies REBEL's WLC.

(a) Assume that REBEL's WLC holds (Eq. 11) for some $\gamma > 0$. By setting $\mathbf{w}_n^+$ and $\mathbf{w}_n^-$ as follows:

$$\mathbf{w}_n^+ = w_n \Big( \frac{1+y_n}{2K} \Big) \mathbf{1}, \;\; \mathbf{w}_n^- = w_n \Big( \frac{1-y_n}{2K} \Big) \mathbf{1} \qquad \therefore \;\; \forall\, w_n \geq 0 \;\; \Rightarrow \;\; \mathbf{w}_n^+, \mathbf{w}_n^- \geq \mathbf{0}$$

then, since REBEL's WLC assumedly holds, it in turn simplifies to:

$$\exists\, f \quad \text{satisfying:} \quad \Big| \sum_{n=1}^{N} w_n y_n f(\mathbf{x}_n) \Big| \geq \gamma \sum_{n=1}^{N} w_n$$
$$\Rightarrow \;\; \exists\, g = f \text{ or } -f \quad \text{satisfying:} \quad \sum_{n=1}^{N} w_n y_n g(\mathbf{x}_n) \geq \gamma \sum_{n=1}^{N} w_n$$

which is exactly the form of the Binary WLC. Note that it is implicitly assumed that if a weak learner $f$ exists, then its trivial negation $-f$ also exists.

(b) Conversely, assume that the Binary WLC holds (Eq. 9) for some $\gamma > 0$. Let $j$ be the class contributing the largest portion of the weight, and set $y_n$ and $w_n$ as follows:

$$j \equiv \arg\max_k \sum_{n=1}^{N} \langle |\mathbf{w}_n^+ - \mathbf{w}_n^-|, \boldsymbol{\delta}_k \rangle \quad \Rightarrow \quad \sum_{n=1}^{N} \langle |\mathbf{w}_n^+ - \mathbf{w}_n^-|, \boldsymbol{\delta}_j \rangle \geq \frac{1}{K} \sum_{n=1}^{N} \langle |\mathbf{w}_n^+ - \mathbf{w}_n^-|, \mathbf{1} \rangle$$

$$y_n = \langle \text{sign}[\mathbf{w}_n^+ - \mathbf{w}_n^-], \boldsymbol{\delta}_j \rangle, \quad w_n = \langle |\mathbf{w}_n^+ - \mathbf{w}_n^-|, \boldsymbol{\delta}_j \rangle \geq 0$$

Since the Binary WLC assumedly holds, it can be expanded as:

$$\exists\, f \quad \text{satisfying:} \quad \Big\langle \sum_{n=1}^{N} [\mathbf{w}_n^+ - \mathbf{w}_n^-] f(\mathbf{x}_n), \boldsymbol{\delta}_j \Big\rangle \geq \gamma \sum_{n=1}^{N} \langle |\mathbf{w}_n^+ - \mathbf{w}_n^-|, \boldsymbol{\delta}_j \rangle \geq \frac{\gamma}{K} \sum_{n=1}^{N} \langle |\mathbf{w}_n^+ - \mathbf{w}_n^-|, \mathbf{1} \rangle$$

Finally, since the sum of absolute values is always positive:

$$\sum_{k \neq j} \Big\langle \Big| \sum_{n=1}^{N} [\mathbf{w}_n^+ - \mathbf{w}_n^-] f(\mathbf{x}_n) \Big|, \boldsymbol{\delta}_k \Big\rangle \geq 0$$

$$\therefore \quad \exists\, f \quad \text{satisfying:} \quad \Big\langle \Big| \sum_{n=1}^{N} [\mathbf{w}_n^+ - \mathbf{w}_n^-] f(\mathbf{x}_n) \Big|, \mathbf{1} \Big\rangle \geq \gamma' \sum_{n=1}^{N} \langle |\mathbf{w}_n^+ - \mathbf{w}_n^-|, \mathbf{1} \rangle$$

which is exactly the form of REBEL's WLC, with $\gamma' \equiv \gamma/K > 0$.

We have just shown that the REBEL and Binary weak learning conditions are equivalent. Therefore, REBEL is Boostable, exponentially reducing the training risk without imposing too-weak or too-strong requirements on its binary weak learners [27].

## S.8 Reduction to Binary

In the binary (2-class) cost-insensitive case, the surrogate loss reduces to:

$$\mathcal{L} = \frac{1}{2N} \sum_{y_n=1} \big( F(-H_{n1}) + F(H_{n2}) \big) + \frac{1}{2N} \sum_{y_n=2} \big( F(H_{n1}) + F(-H_{n2}) \big) \tag{12}$$

$$\text{where for brevity, we denote:} \quad \sum_{y_n=k} (\ldots) \equiv \sum_{n=1}^{N} \mathbb{1}(y_n = k)(\ldots) \qquad \text{(i.e. sum over data in class } k\text{)}$$

and: $H_{n1} \equiv \langle \mathbf{H}_T(\mathbf{x}_n), \boldsymbol{\delta}_1 \rangle,\ H_{n2} \equiv \langle \mathbf{H}_T(\mathbf{x}_n), \boldsymbol{\delta}_2 \rangle,\ H_{n1}^+ \equiv \langle \mathbf{H}_{T+1}(\mathbf{x}_n), \boldsymbol{\delta}_1 \rangle,\ H_{n2}^+ \equiv \langle \mathbf{H}_{T+1}(\mathbf{x}_n), \boldsymbol{\delta}_2 \rangle$

**Lemma**: $H_{n1} = -H_{n2}$

**Proof**: (by induction)

Before any training: $\mathbf{H}_0(\mathbf{x}_n) \equiv \mathbf{0} \;\Rightarrow\; H_{n1} = H_{n2} = 0 \;\Rightarrow\; H_{n1} = -H_{n2} \;\; \forall n$

Assume this holds for $\mathbf{H}_T(\mathbf{x})$. On the $T{+}1^{\text{th}}$ iteration, $\mathbf{H}_{T+1}(\mathbf{x}) \equiv \mathbf{H}_T(\mathbf{x}) + f(\mathbf{x})\,\mathbf{a}$ hence:

$$\begin{aligned}
\mathcal{L}_{T+1} &= \frac{1}{2N} \Big( \sum_{y_n=1} F(-H_{n1}^+) + F(H_{n2}^+) + \sum_{y_n=2} F(H_{n1}^+) + F(-H_{n2}^+) \Big) \\
&= \frac{1}{2N} \sum_{y_n=1} \big( F(-H_{n1} - f(\mathbf{x}_n) a_1) + F(H_{n2} + f(\mathbf{x}_n) a_2) \big) + \frac{1}{2N} \sum_{y_n=2} \big( F(H_{n1} + f(\mathbf{x}_n) a_1) + F(-H_{n2} - f(\mathbf{x}_n) a_2) \big)
\end{aligned}$$

Hence, solving for optimal $a_1$ by setting: $\frac{\partial \mathcal{L}_{T+1}}{\partial a_1} = 0$

$$\therefore \sum_{y_n=1} f(\mathbf{x}_n) F'(-H_{n1} - f(\mathbf{x}_n) a_1) = \sum_{y_n=2} f(\mathbf{x}_n) F'(H_{n1} + f(\mathbf{x}_n) a_1) \tag{13}$$

Whereas solving for optimal $a_2$ by setting: $\frac{\partial \mathcal{L}_{T+1}}{\partial a_2} = 0$

$$\therefore \sum_{y_n=1} f(\mathbf{x}_n) F'(H_{n2} + f(\mathbf{x}_n) a_2) = \sum_{y_n=2} f(\mathbf{x}_n) F'(-H_{n2} - f(\mathbf{x}_n) a_2)$$

Since by assumption, $H_{n1} = -H_{n2}$, therefore:

$$\therefore \sum_{y_n=1} f(\mathbf{x}_n) F'(-H_{n1} + f(\mathbf{x}_n) a_2) = \sum_{y_n=2} f(\mathbf{x}_n) F'(H_{n1} - f(\mathbf{x}_n) a_2) \tag{14}$$

Note that if $a_1 = a$ solves Eq. 13, then $a_2 = -a$ solves Eq. 14; hence: $a_1 = -a_2$. Therefore:

$$H_{n1}^{+} = H_{n1} + f(\mathbf{x}_n) a_1 = -H_{n2} - f(\mathbf{x}_n) a_2 = -(H_{n2} + f(\mathbf{x}_n) a_2) = -H_{n2}^{+}$$

On the $T{+}1^{\text{th}}$ iteration, again: $H_{n1}^{+} = -H_{n2}^{+}$

Back to the original binary loss (Eq. 12):

$$\mathcal{L} = \frac{1}{2N} \sum_{y_n=1} \big( F(-H_{n1}) + F(H_{n2}) \big) + \frac{1}{2N} \sum_{y_n=2} \big( F(H_{n1}) + F(-H_{n2}) \big)$$

$$= \frac{1}{N} \Big( \sum_{y_n=1} F(-H_{n1}) + \sum_{y_n=2} F(H_{n1}) \Big) = \frac{1}{N} \sum_{n=1}^{N} F(-y_n^\star H_{n1})$$

$$\text{where: } y_n^\star \in \{+1, -1\} \text{ for: } y_n \in \{1, 2\}$$

which is exactly the same form as the loss function for binary boosting. For instance, if $F(x) \equiv e^x$, REBEL reduces to AdaBoost, $F(x) \equiv \log_2(1 + 2^x)$ reduces to LogitBoost, and $F(x) \equiv (1 + x)^2$ reduces to $L_2$-Boost.

## S.9 Cost-Sensitive Training

Fig. 7 demonstrates the behavior of classifiers trained using four different cost matrices: (a) uniform misclassification costs, (b) $10\times$ higher green false-negative costs, (c) $10\times$ higher red false-positive costs, and (d) $10\times$ higher red false-negative costs. As expected, REBEL adapts its behavior according to the input misclassification costs. Note from Fig. 7 that REBEL avoids missing a class with high false negative cost at the expense of missing other classes.

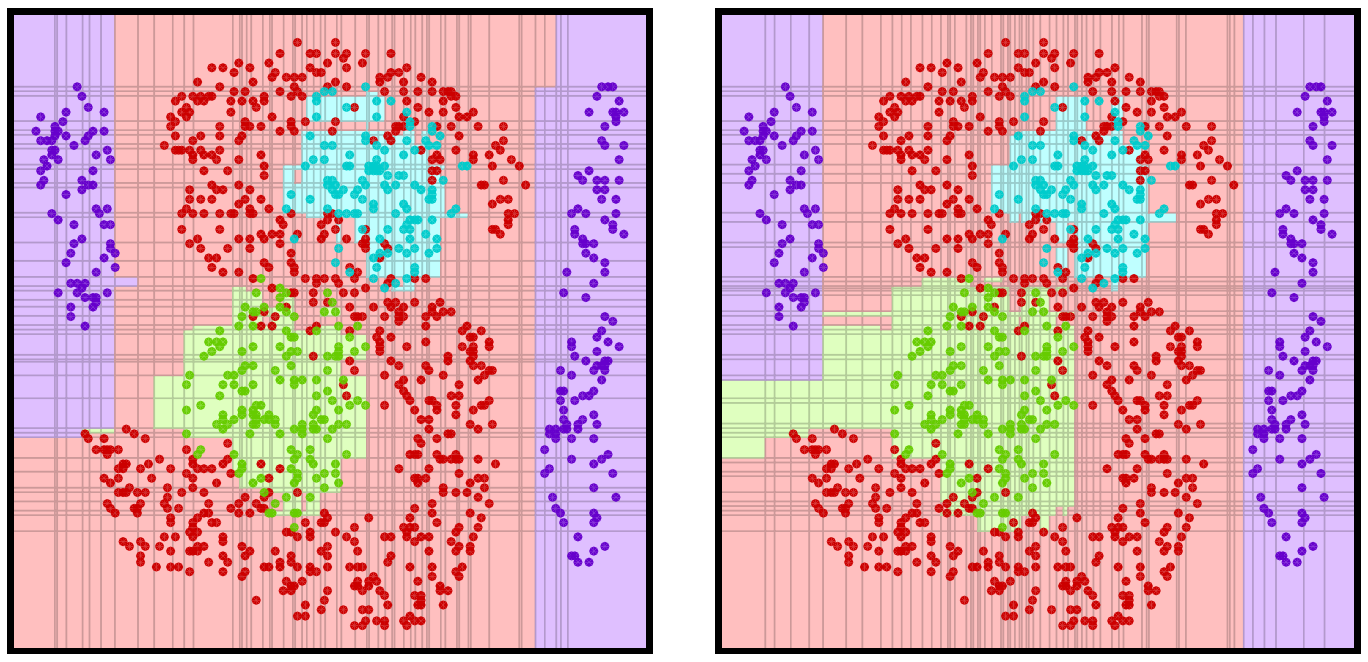

(a) Uniform misclass. costs (b) High green false-neg. costs

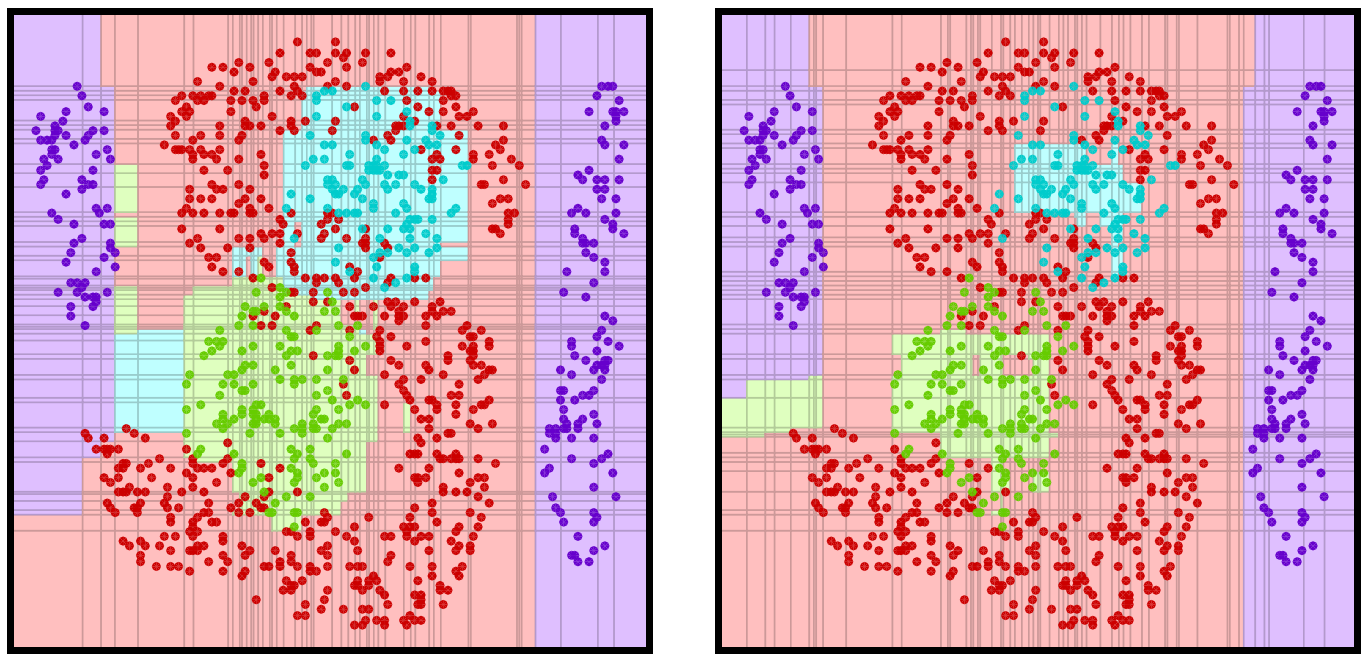

(c) High red false-pos. costs (d) High red false-neg. costs

Figure 7: Effects of cost-sensitive training on a synthetic dataset, trained using REBEL with 100 depth-2 trees. Classes are color-coded, and each panel is the result of a different cost matrix. (c) and (d) demonstrate the difference in output with a high false-positive cost versus a high false-negative cost. As encoded in the cost matrix, in (c), no red classification is incorrect, whereas in (d), no red datapoint is misclassified.

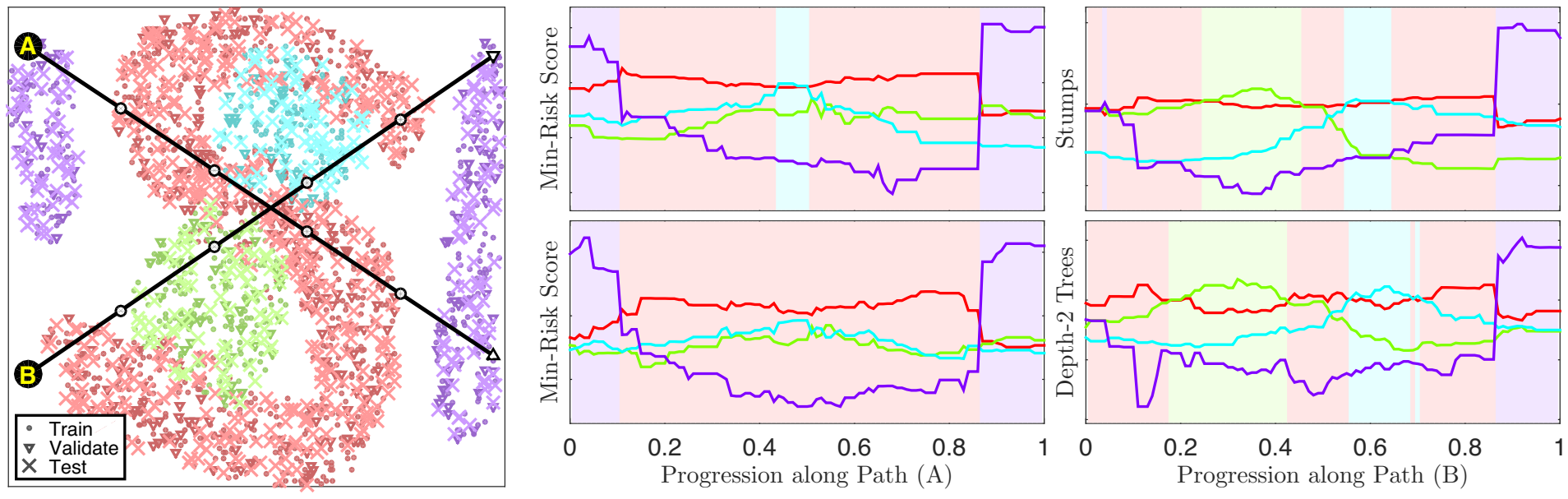

Figure 8: REBEL output on a synthetic 2D dataset, sampled along two paths labeled (A) and (B). Nibs along the two paths on the left-most panel are at 20% increments. Two classifiers were trained: the first using 100 stumps (top), the second using 100 depth-2 trees (bottom). Min-risk scores (i.e. $\mathbf{H}(\mathbf{x})$) are plotted (the background colors correspond to the minimum-risk class).

Given an input, the output of REBEL is an estimate of the corresponding minimum-risk class. Fig. 8 shows the "min-risk score" (i.e. $\mathbf{H}(\mathbf{x})$) over two paths through the input feature space, using either stumps or depth-2 trees as weak learners. The scores vary across class boundaries while staying fairly constant within class-specific regions.

## S.10 Synthetic Random Datasets

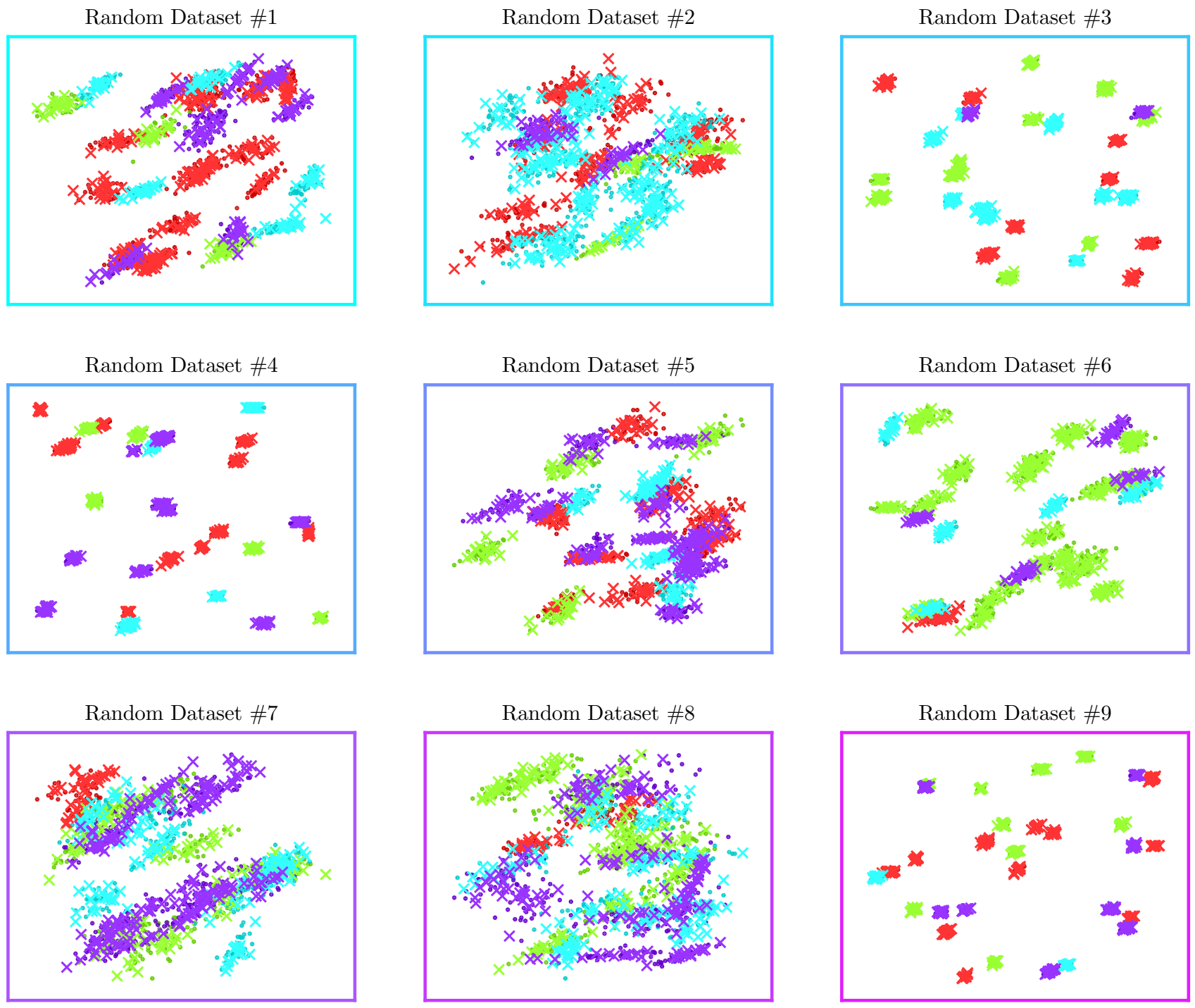

Figure 9: Random datasets generated for Fig.5

## S.11 Pictorial results of SUN and CUB-200

Figure 11 shows some visual examples of REBEL's classification on SUN. Due to the nature of the features used (HOG), classification on structured indoor scenes (e.g. Library) is generally more accurate than outdoor scenes. Figure 10 shows some visual examples of REBEL's classification on the CUB-200 *Woodpecker Vireo* subset. The richness of neural-net based features combined with REBEL cost-sensitive training leads to very few costly mistakes.

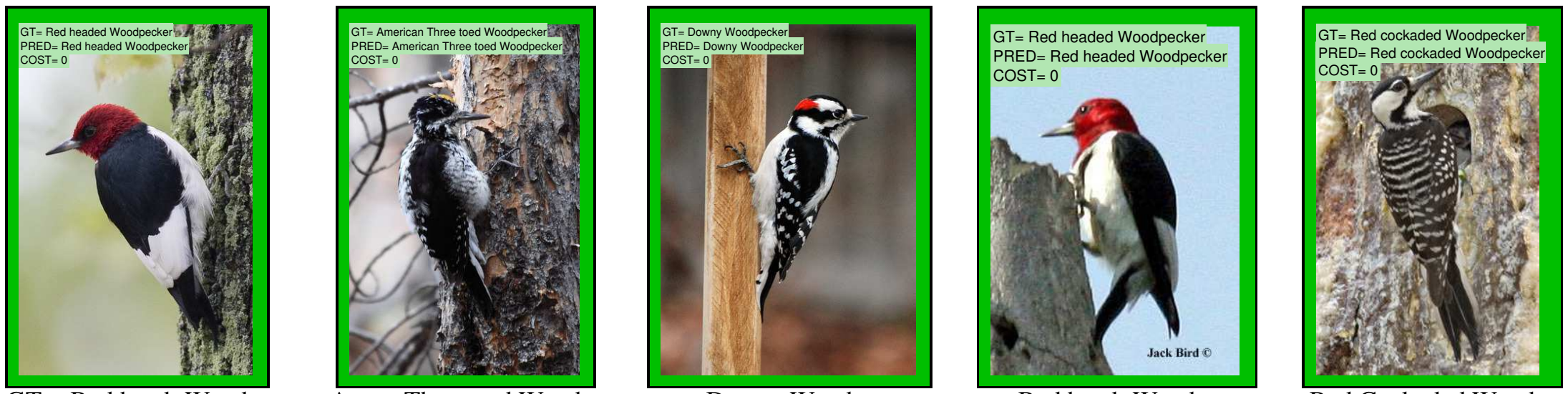

GT = Red head. Woodp. PRED = Red head. Woodp. COST = 0

Amer. Threetoed Woodp. Amer. Threetoed Woodp. COST = 0

Downy Woodp. Downy Woodp. COST = 0

Red head. Woodp. Red head. Woodp. COST = 0

Red Cockaded Woodp. Red Cockaded Woodp. COST = 0

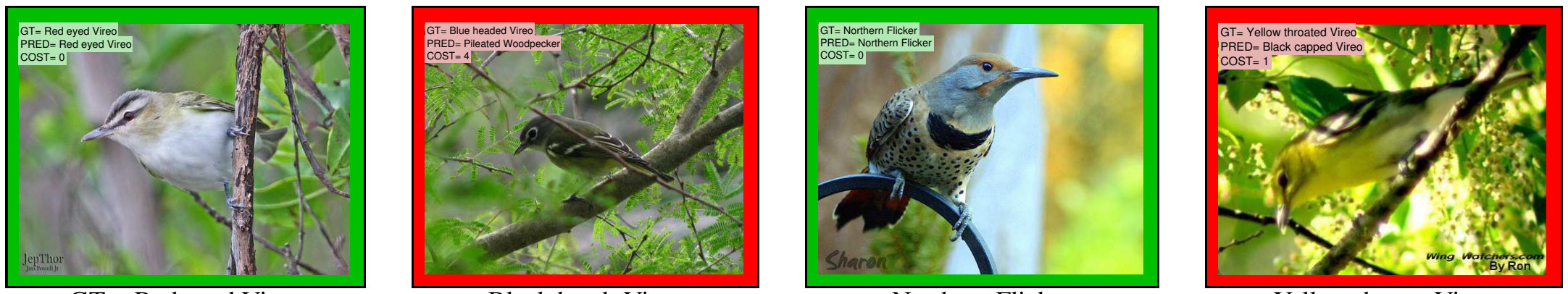

GT = Red eyed Vireo PRED = Red eyed Vireo COST = 0

Black head. Vireo Pileated Woodp. COST = 4

Northern Flicker Northern Flicker COST = 0

Yellow throat. Vireo Black cap. Vireo COST = 1

Figure 10: REBEL visual examples of CUB birds classification. First row: 5 lowest-risk examples on test set; second row: 4 highest-risk examples. Costly mistakes happen very rarely due to REBEL's use of taxonomic costs. (GT means ground truth, PRED is prediction)

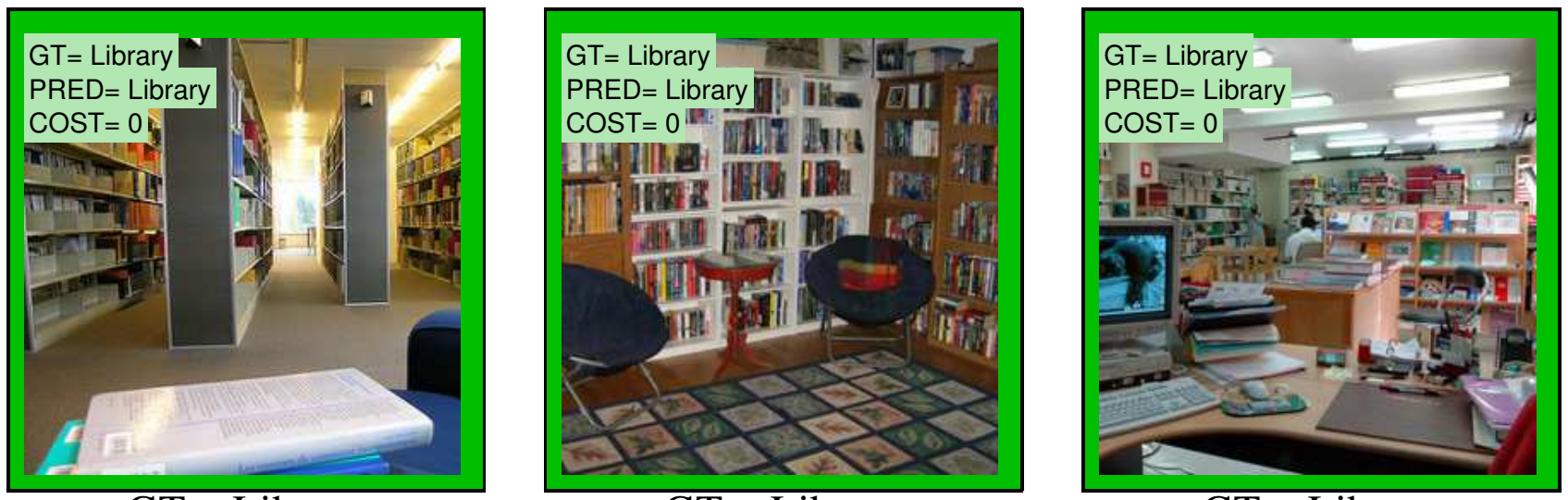

GT = Library PRED = Library COST = 0

GT = Library PRED = Library COST = 0

GT = Library PRED = Library COST = 0

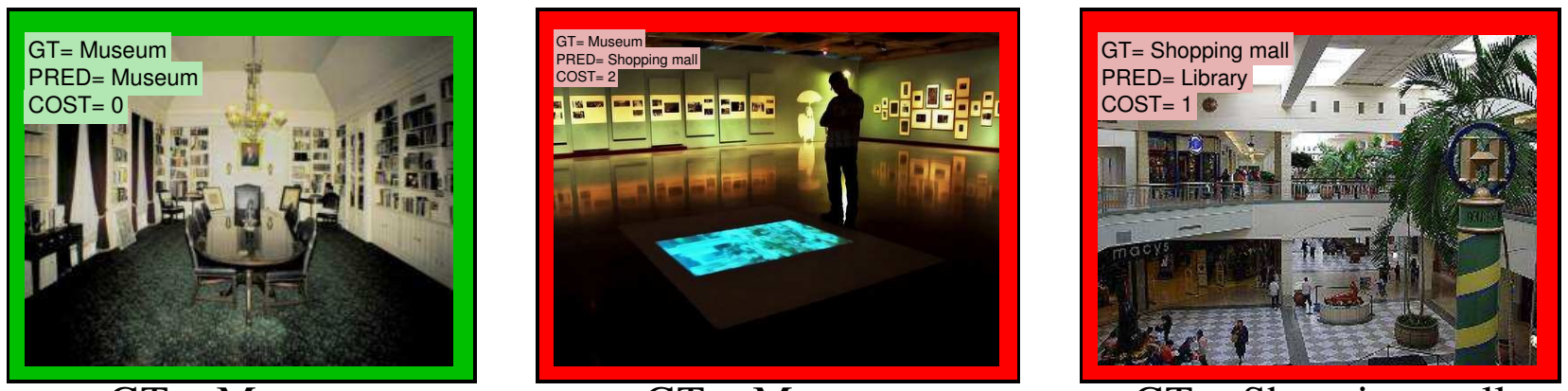

GT = Museum PRED = Museum COST = 0

GT = Museum PRED = Shopping mall COST = 1

GT = Shopping mall PRED = Library COST = 1

Figure 11: REBEL visual examples of SUN scenes classification. First row: three lowest-risk examples; second row: three highest-risk examples. (GT means ground truth, PRED is prediction)